# Modeling Neural Architecture Search Methods for Deep Networks

Emad Malekhosseini, Mohsen Hajabdollahi, Nader Karimi, Shadrokh Samavi

Department of Electrical and Computer Engineering
Isfahan University of Technology Isfahan 84156-83111, Iran

*Abstract*— **There are many research works on the designing of architectures for the deep neural networks (DNN), which are named neural architecture search (NAS) methods. Although there are many automatic and manual techniques for NAS problems, there is no unifying model in which these NAS methods can be explored and compared. In this paper, we propose a general abstraction model for NAS methods. By using the proposed framework, it is possible to compare different design approaches for categorizing and identifying critical areas of interest in designing DNN architectures. Also, under this framework, different methods in the NAS area are summarized; hence a better view of their advantages and disadvantages is possible.**

*Keywords*— ***automatic deep neural network design, neural architecture search, efficient network structure, modeling***

## I. Introduction

A lot of research has been done on improving the viability of using deep neural networks (DNNs) for different applications, all of which depend on either increasing the DNN accuracy or decreasing DNN costs such as memory and computation. For example, in many realtime applications, it is essential to minimize the size of the architecture of a DNN [1-4]. Among various architectural improvement researches, neural architecture search (NAS) methods can be considered as an effective approach. Architectural decisions and strategies in designing DNNs are summarized under the NAS methods that lead to a target architecture for DNNs with defined constraints.

The goal of NAS methods is to derive an optimal architecture for an application given a set of constraints, by incrementally searching for better architectures that satisfy those constraints. Setting an architecture search space and configuring its parameters, determines the *N*-dimensional architecture space containing all of the possible architectures for an application. Each structure can be represented by an *N*-tuple where *N* is the number of architecture search space parameters.

To design an efficient architecture that meets all of the problem's constraints, a set of cost terms needs to be assigned to those constraints. Searching under the given constraints narrows the search space by excluding the high-cost architectures and including the low-cost ones. Different studies in the NAS area can be categorized based on the manual and automatic searching methods.

In the area of manual searching methods, in [4], neural elements are modified in the convolutional neural network (CNN) by imposing sparsity constraints on the objective function during training. In [6], the 3×3 convolutional building blocks are replaced with a building block comprising of a combination of 3×3 and 1×1 convolutional operators in such a way that the number of input channels of each building block is reduced. Also, down-sampling is applied late to keep the feature map sizes large and to increase the classification accuracy. In [7], building block level modification is employed in which one building block is progressively added to the network to make a cascaded structure. In [8], operator level modification is applied to the building blocks of a CNN by introducing dense skip connections between the convolutional operators in a building block. In [9] A pre-trained low-level feature extraction sub-network is used, which is shared between several high-level feature extraction and classification sub-networks. In [10], sub-network level modification is proposed in which, based on the class hierarchies in the dataset, the proposed architecture divides the DNN into several sub-networks into several layers. The first layer sub-network is a shared low-level feature extractor that is shared between the second and third layers, and the sub-network in the second layer is a coarse-grained classifier that predicts the general classes. The sub-network in the third layer is comprised of fine-grained classifier sub-networks. In [11], the same idea as [10] is pursued. Also, in [11], a sub-network level modification is employed to decrease the inference time. Several sub-networks are specialized for each set of input classes and only run those sub-networks for those classes. To increase the facial key-point detection accuracy, Sun et al. [12], offer a building-block level modification, and a cascading architecture with three levels is proposed. In each level, there are several groups of building blocks where each group is dedicated to the detection of a specific facial key-point. Conventional NAS methods are manual and require immense knowledge and experience about the DNN structures. Also, there is a cycle of trials and errors to find the optimal architecture for a given application. Automatic NAS methods are hence developed to automate the design process by parameterizing the network architecture and searching a suitable architecture for a given use. The automated approach is referred to as neural architecture auto-search [13].

Auto-search methods in the NAS area can be defined as a problem to find a model $a^*$ in the architecture search space $A$ that when trained on a training dataset $d_{train}$ minimizes a loss function L as Eq. (1). Also, a constraint should be set about the validation set which is maximizing an objective function O on the inference testing dataset $d_{valid}$.

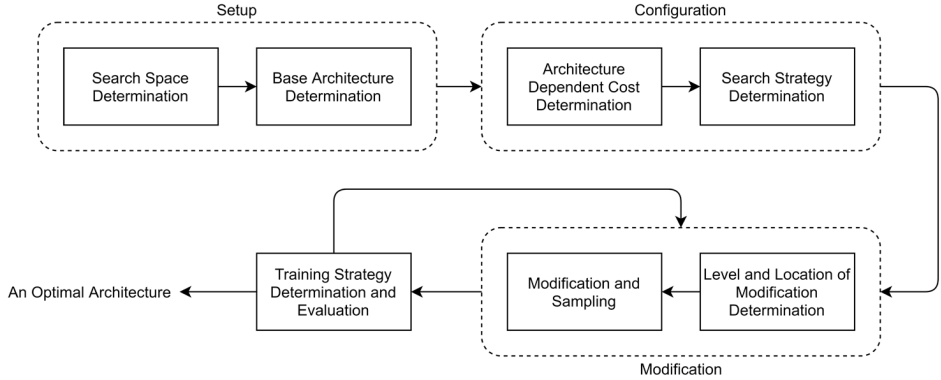

Fig 1. General structure of the proposed modeling for methods in NAS area

In previous researches in the area of auto-search, Liu et al. [13] found a high-performance architecture for dense image prediction. A hierarchical search space is considered including two spaces, micro-architecture search space, and the macro-architecture search space. In [14], a method is proposed to automatically fine-tune the building-blocks and connections of a CNN using building block modifications. Tan et al. [15] proposed a reinforcement learning-based auto-search method to design architecture. In each iteration, a controller samples the architecture that maximizes the objective function. In [16], a direct acyclic graph (DAG) containing multiple sub-graphs is considered. Each sub-graph is in the form of a smaller direct acyclic graph, which realizes a sample architecture. In [17], an auto-search method is proposed in which the search space is hierarchical with fixed macro-architecture and is represented by a stochastic super-network. Inspired by the transfer learning methods, Zoph et al. proposed a reinforcement learning-based auto-search method [18]. In their approach, the search space has a fixed macro-architecture comprising of normal cell and reduction cell micro-architectures. Real et al. [19] use the search space described in [18] and applies a novel evolutionary algorithm as the search method.

Although there are numerous methods for automated architecture design and guidelines to guide the manual architecture design, to the best of our knowledge, there is no unifying model in which these methods can be embedded, explored, and compared. Our goal in this paper is to propose such a unifying DNN architecture design model. Under the proposed model, all the studies can be methodically explored and compared, their strengths and weaknesses can be identified and directions for improvements suggested.

As illustrated in Fig. 1, the proposed model constitutes of four major stages. The first stage is the setup phase in which the search space and starting baseline architecture are determined and initialized. The second stage is the configuration phase in which the architecture search strategies and guidelines and a way to measure the cost of an existing architecture are determined. The third stage is a modification, which, based on the decisions made in the previous step, chooses the level and location of the needed modifications to the architecture. The fourth stage is architecture evaluation, in which the training strategy for the next stage is determined. If the performance is satisfactory, the search is stopped. The third and fourth stages form a loop in which iterative improvements are applied to the architecture. In the following of the current study, the details of each step are presented.

The remaining of this paper is organized as follows. In sections II and III, the setup and configuration for NAS methods are presented respectively. In Section IV, and V, the modification, and evaluation stages are explained respectively. In section VI, the modeling of different methods is presented. Finally, section VII is dedicated to the concluding remarks of this study.

II. SETUP PHASE

The setup stage is the first stage of the framework, which includes two main steps, as illustrated in Fig. 1. These steps are explained in the followings section.

*A. Search space determination*

The determination of the architecture search space is mostly affected by the target application because different applications impose different performance constraints on the architecture under design. The area in which the architecture is searched is affected by the available computational resources. There is a trade-off between the flexibility and the generality of a search space and the computational requirements of the searching operation. Thus, an efficient configuration of the search space to balance the flexibility and the computational requirements is very important. The architecture search space need only be strictly defined in auto-search methods and is usually not intended for manual search methods. Search spaces can be explored through four abstraction levels.

- Operator level
- Connection level
- Cell level
- Building block level
- Sub-network level

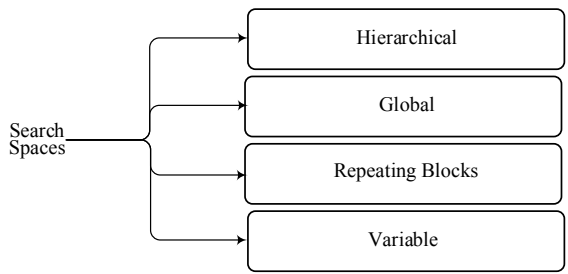

Fig. 2. Search spaces in NAS methods

A cell is defined as a small convolutional block. Several cells connected in various ways form a building block. A sub-network is a small network that performs a specific task, e.g., a low-level feature extractor, and is comprised of several building blocks connected in various ways.

Generally, search spaces can be categorized into the following categories, as illustrated in Fig. 2.
- Hierarchical
- Repeating Blocks
- Global
- Variable

Hierarchical search spaces are typically comprised of two search levels, micro-architecture search space, and macro-architecture search space, where the macro-architecture specifies how the microarchitectures are connected to make the entire network. Hierarchical search spaces can employ either a fixed micro-architecture or a fixed micro-architecture to make the search space more manageable. Hierarchical search spaces are employed in [12][13][14][15][17]. The search space in [14] is in the form of a fixed micro-architecture and, in [15][17], is in the form of a fixed macro-architecture.

In the repeating blocks search spaces, similar blocks possibly at different abstraction levels, i.e., cells, building blocks, and sub-networks are repeatedly attached to create the final network [6][8][18][19].

In the global search spaces, the entire scope of the baseline network is searched, and area under exploration is not limited to an entity from an architectural point of view [5][10][11][16]

Variable search spaces are used when the search space needs to be changed; because the requirements of the application are not known beforehand [7][9].

*B. Baseline architecture determination*

After the determination of architecture search space, a base architecture working under the defined search space needs to be determined. This architecture is used as a baseline to experiment with different network architectures in both manual and automatic searching methods. Baseline architectures can be in the form of the following categories.
- Super-Network
- Direct acyclic graph (DAG)
- Randomly chosen architecture in the search space
- Given architecture
- Module

A super-network is defined as the superposition of all possible architecture within the search space [13][14][17].

Sub-graphs of a direct acyclic graph correspond to different possible architectures within the search space, i.e., in a DAG, each edge corresponds to an operation such as convolution or down-sampling [16]. Usually, when reinforcement learning or evolutionary algorithms are used, a random architecture within the search space is taken as the baseline architecture [15][18][19].

In some cases, the architecture search is done by searching for a better architecture given an arbitrary architecture like Res-Net or VGG [5][9][11]. Sometimes a convolutional module is taken as the baseline architecture, and the architecture search is done through extending or repeating the baseline module in various ways [6][7][8][10][12].

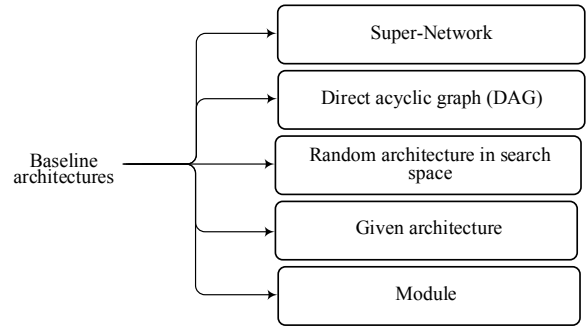

Fig. 3. Baseline architecture

III. CONFIGURATION

This stage plays an essential role in designing an efficient architecture for DNNs. In this stage, a way to measure the cost of architecture should be considered. Also, the search strategy suitable for a given application is specified in this stage.

*A. Architecture dependent costs*

Overall, the architecture cost is significant for searching for new architectures. As illustrated in Fig. 4, the overall cost of an architecture is defined based on different cost terms such as inference latency [15][17], training time [9][13], complexity [14][17], computation [7][16][18], memory consumption [7][16], energy [9], search cost [13] and training time [9]. In Fig. 4, the corresponding ways to realize the defined costs are illustrated. The cost of a NAS method is used to guide the search strategy in the subsequent stages. Also, the determination of architecture-dependent cost metrics provides a way to compare different architectures and can be used to guide the intuition behind the search strategy.

*B. Search strategy determination*

The search strategy is the procedure through which the search space is explored to find the best architecture. Usually, the search space of all possible architectures can be huge, and hence it is not feasible to search all of the architectures existing in that search space. The decisions made in this step are dependent on the selection of the baseline architecture and the search space. Also, the search strategy is dependent on the overall architecture cost, which is used to compare the architectures in the search space. As illustrated in Fig. 5, several automatic and manual search methods are used to design new architectures.

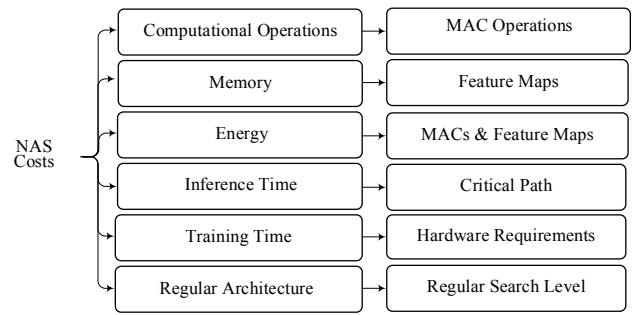

Fig. 4. NAS costs and corresponding ways to their realization

In the manual search, the search strategy can be in the form of some guidelines or rules for designing an architecture. One such rule or search strategy can be the decision to add the architectural components in forms such as cascade or sequential as in [6][7][12]. Another search strategy can be to search for dense architecture by introducing skip connections as in [8] or to search for compact architectures as in [5]. Also, a search strategy can be to search for bifurcated architectures that increase feature reuse and parameter sharing as in [9][10][11]. Furthermore, another search strategy is to search for an ensemble of networks and use the average of their outputs to produce the final result [10][11][12].

As illustrated in Fig. 5, in auto-search based methods, depending on the selected baseline architecture, one method for the search strategy is differentiable architecture search through which using gradient descent methods a super-network or a DAG is trained to minimize an overall architecture cost function [13][14][16][17]. Another search strategy is to define a reward based on reducing the total architecture cost and use reinforcement learning algorithms to learn better network configurations [15][18]. There are also evolutionary search strategies [19] and random search which achieve comparable results to the other search strategies [20][21].

## IV. MODIFICATION STAGE

The third stage of the proposed framework is the modification phase. This stage and the fourth stage form a loop that incrementally improves the predicted architecture by iteratively modifying the current architecture. This stage constitutes two following steps named "level and location of the modification" and "modification and sampling."

### A. Level and location of the modifications

In this step, the level, location, and method of modification are determined. There are different levels and positions in NAS methods, as shown in Fig. 6. In manual search strategies, the level of modifications can be in forms such as operator level, as stated in [5][8], building block level as in [7][9][12] sub-network level or a combination of them [10][11]. In manual search methods, the modifications are usually determined by intuition, experiment, and designer experience. The location of modifications can be global [10][11], layer-wise [7][8][9][12], and group-based [5][12]. The choice depends on the desired architecture and the given application.

Modifications in an abstraction level are applied by adding or removing the abstract entities on that level. For example, operator level modification can be applied by removing certain operators [5], cell level modification can be applied by replacing the original cell by a smaller cell as presented in [6] or a densely connected cell as in [8]. In automated search methods, the level and location of modification are determined by the parametrized architecture search space. The actual level and the location of the needed modifications are determined by the search strategy acting upon the search space. Depending on the search space, modification levels can be at the operator level [6][13][15][16][17][18][19], cell level [13] and building block level [14][16] and the location can be global [6][14][18][19] and etc. Also, different modification levels may be applied as hybrid levels [13].

### B. Modification and Sampling

In this step, the architecture configuration parameters are learned. In automated search methods, applying the changes results in an architecture configuration distribution. This distribution leads to architectures that best match the set of requirements defined based on the target application. The learned distribution must be sampled to produce a matching architecture. The sampling methods are very different and mostly dependent on the implementation of the search space and the search strategy, but most of the sampling methods can be regarded as a simple pruning to remove unnecessary components from the architectures. In general, sampling methods can be in the form of the categories illustrated in Fig. 7.

Distribution based sampling is employed in [17], and thresholding a super-network is used in [14][5]. In [5], the given architecture can be interpreted as a super-network containing the super-position of all possible sparse architectures. Sampling by the pathfinding through a DAG is used in [13][16], and sampling by Softmax component prediction is used in [18][15].

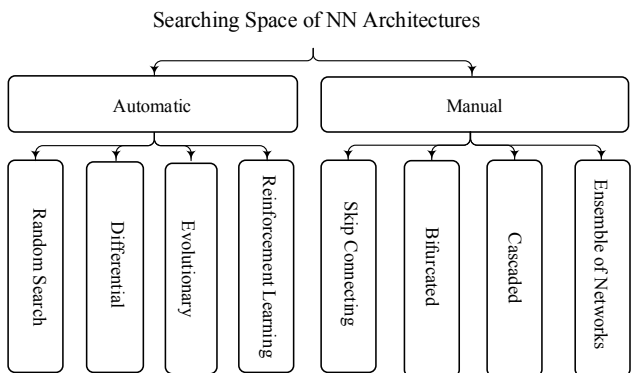
Fig. 5. Architecture search methods in NAS

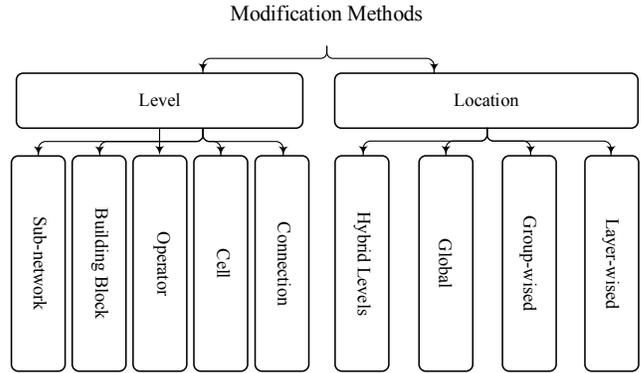
Fig. 6. Modification Methods in NAS

Sampling procedures are not required in the manual search methods as there is no parameterized architecture distribution to be learned.

## V. Architecture Evaluation

In this stage, the sampled architecture is trained and evaluated. Depending on the architecture and the target application, a different training strategy can be used. Some other training strategies include progressive training, where only the newly added modifications to the network are trained while the rest of the network parameters are fixed [7][9]. In the bifurcated training, several sub-networks that are connected to a shared sub-network are trained in parallel [10][11]. Also, the combination of several training strategies as training steps is possible and could increase the accuracy, as stated [7][10]. After training, the architecture's performance is evaluated using the validation set of the dataset. If the performance requirements for the given application are satisfied, the search is terminated and the current architecture is considered as the result. Otherwise, modifications need to be done in order to find architectures with higher performances, and the design flow loops back to the modification stage.

## VI. Modeling of different methods

Using the proposed method, it is possible to summarize various processes under the model. In Table I and Table II, multiple studies in the area of automatic NAS and manual NAS are outlined under the proposed model, respectively. The summarization result is based on the essential aspects of NAS methods. Among different methods and techniques employed in the manual NAS, it can be identified that non-flat structures such as bifurcated and cascaded are widely adopted. Also, building block modifications are utilized for modification. In the automatic NAS methods, search spaces are concentrating on the differential algorithms. Also, search spaces are considered in the form of a cascaded structure.

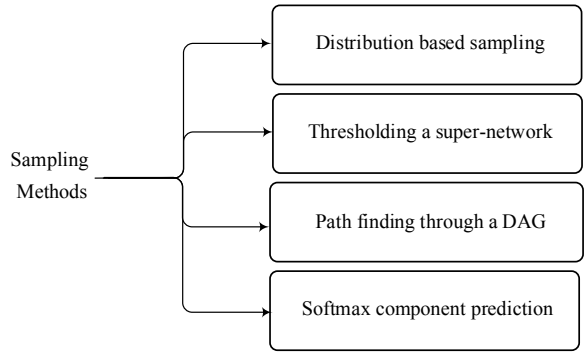

Fig. 7. Various existing sampling methods.

From Table I and Table II, it is possible to say that determining the following components has an essential role in having an efficient architecture.
- Architecture search space and the baseline architecture.
- Architecture search strategy or architecture modification strategy.
- Architecture-dependent cost evaluation and performance metrics.

## VII. Concluding remarks

New modeling for methods and techniques within the scope of NAS was proposed. In the proposed model, all of the critical aspects of the various techniques in the NAS area was covered. A wide diversity was observed in the methods and techniques used in the NAS area, which necessitates having a general model to summarize different routines. Various approaches were explored and summarized under the proposed model. With the proposed model, achieving better insights into the most important and recent contributions conducted in designing NAS methods is made possible. Also, the proposed model can be further extended by the future researcher to act as a guideline for designing new methods in NAS.

TABLE I. Summarizing different techniques in manual NAS based on the proposed model

| Work | Cost | Baseline Architecture | Search Space | Search Algorithm | Modification Method | Sampling Method | Dataset |
|---|---|---|---|---|---|---|---|
| [6] | Parameters | Module | Repeating blocks | Cascade | Operator Global | - | Image-Net |
| [7] | Computation Memory | Module | Variable | Cascade | Building block Layer-wise | - | CIFAR-10 CIFAR-100 |
| [8] | Task Error | Module | Repeating blocks | Skip connecting | Connection Layer-wise | - | CIFAR-10 Image-Net |
| [9] | Training Time | Given architecture | Variable | Bifurcate | Building block Layer-wise | - | CIFAR-10 Image-Net |
| [10] | Task Error | Module | Global | Bifurcate Ensemble | Subnetwork Global | - | CIFAR-100 Image-Net |
| [12] | Task Error | Module | Hierarchical | Cascade Ensemble | Building block Layer-wise | - | BioID LFPW |
| [11] | Task Error Inference efficiency | Given architecture | Global | Bifurcate Ensemble | Sub-network Global | - | CIFAR-100 Image-Net |

TABLE II. Summarizing different techniques in automatic NAS based on the proposed model.

| Work | Cost | Baseline Architecture | Search Space | Search Algorithm | Modification Method | Sampling Method | Dataset |
|---|---|---|---|---|---|---|---|
| [5] | Parameters | Given architecture | Global | Differential | Operator Level Group-wise Layer-wise | Thresholding | MNIST CIFAR-10 Image-Net |
| [13] | Search cost | Super-network | Hierarchical | Differential | Operator, Cell Hybrid | Path finding | Cityscapes |
| [14] | Complexity | Super-network | Hierarchical fixed micro | Differential | Building block Global | Thresholding | SVHN VGG-Flowers |
| [15] | Latency | Random architecture in search space | Hierarchical fixed macro | Reinforcement learning | Operator Layer-wise | Softmax | Image-Net |
| [16] | Arbitrary Computation Memory | Direct acyclic graph | Global | Differential | Operator Building block Layer-wise | Path finding | CIFAR-10 CIFAR-100 |
| [17] | Latency Size | Super-network | Hierarchical fixed macro | Differential | Operator Layer-wise | Distribution sampling | Image-Net |
| [18] | Computation Task error | Random architecture in search space | Repeating block | Reinforcement learning | Operator Global | Softmax | CIFAR-10 Image-Net |
| [19] | Task error | Random architecture in search space | Repeating block | Evolutionary | Operator Global | - | CIFAR-10 Image-Net |